\documentclass[conference]{IEEEtran}
\IEEEoverridecommandlockouts
\usepackage{tabularx}
\usepackage{multirow}
\usepackage{booktabs}
\usepackage{amsmath,amssymb,amsfonts}
\usepackage{algorithmic}
\usepackage{graphicx}
\usepackage{hyperref}
\usepackage{textcomp}
\usepackage{xcolor}
\usepackage[backend=bibtex]{biblatex}
\begin{document}

\title{GANji: A Framework for Introductory AI Image Generation}

\author{\IEEEauthorblockN{1\textsuperscript{st} Chandon Hamel}
\IEEEauthorblockA{\textit{Anderson College of Business and Computing} \\
\textit{Regis University}\\
Denver, Colorado \\
chamel@regis.edu}
~\\
\and
\IEEEauthorblockN{2\textsuperscript{nd} Mike Busch}
\IEEEauthorblockA{\textit{Anderson College of Business and Computing} \\
\textit{Regis University}\\
Denver, Colorado \\
mbusch@regis.edu}
}

\maketitle

\begin{abstract}
The comparative study of generative models often requires significant computational resources, creating a barrier for researchers and practitioners. This paper introduces GANji, a lightweight framework for benchmarking foundational AI image generation techniques using a dataset of 10,314 Japanese Kanji characters. It systematically compares the performance of a Variational Autoencoder (VAE), a Generative Adversarial Network (GAN), and a Denoising Diffusion Probabilistic Model (DDPM). The results demonstrate that while the DDPM achieves the highest image fidelity, with a Fr\'echet Inception Distance (FID) score of 26.2, its sampling time is over 2,000 times slower than the other models. The GANji framework is an effective and accessible tool for revealing the fundamental trade-offs between model architecture, computational cost, and visual quality, making it ideal for both educational and research purposes.
\end{abstract}


\section{Introduction}

In the quest to understand and implement AI image generation techniques at a low computational cost, the choice of dataset significantly impacts both the feasibility and effectiveness of experimentation. Kanji characters, with their inherent logical structure and visual complexity, serve as an excellent medium for this purpose. Despite being represented as low-resolution, single-channel grayscale images, Kanji possess sufficient detail to effectively train and evaluate generative AI models. Their varied yet recognizable structures offer a straightforward means of qualitatively assessing model performance, allowing anyone with reasonable exposure to Japanese or Chinese writing to distinguish high-quality generated samples. This paper introduces \textbf{GANji}, a framework designed to serve as a ``Hello, World'' for researchers and practitioners entering the field of generative imaging, allowing for meaningful model comparison without prohibitive computational requirements.

Building on this premise, this study compares and analyzes the ability of three foundational generative models to produce Kanji-like images: the Variational AutoEncoder (VAE), the Generative Adversarial Network (GAN), and the Denoising Diffusion Probabilistic Model (DDPM). These models were trained on a dataset of 10,314 Kanji characters to evaluate the impact of hyperparameter adjustments and architectural modifications, thereby comparing the relative strengths and weaknesses of each technique. The emergence of coherent Kanji-like structures in sampled outputs served as a continuous indicator of model performance during training, and the quality of the final generated images could be readily assessed. Ultimately, this work provides both a methodology for comparing image generation models under strict resource constraints and a direct comparison of these models' capabilities on this unique and challenging dataset.

\section{Methodology}

\subsection{Tools}
\subsubsection{Hardware}
Experiments were conducted on a personal machine running Ubuntu 22.04 LTS and equipped with an NVIDIA RTX 4070 GPU (12 GB VRAM). This hardware was selected to demonstrate the framework's feasibility on consumer-grade equipment, which offers significant computational power while having known limitations compared to industrial-grade accelerators.

\subsubsection{Software Environment}
The project was implemented in Python 3.9. Model architectures were defined using PyTorch 2.0, with PyTorch Lightning employed for managing training loops and hardware acceleration. Additional libraries included NumPy for numerical operations and Matplotlib for visualization. To evaluate model performance, Fr\'echet Inception Distance (FID) scores were computed using the \texttt{clean-fid} Python package \cite{parmar2021cleanfid}.

\subsubsection{Version Control and Reproducibility}
To ensure reproducibility, the project was managed with Git for version control. The complete source code is publicly available on GitHub at: \url{https://github.com/Feebami/GANji}.

\subsubsection{Data Management}
The dataset was stored locally. Data loading was handled by the PyTorch \texttt{DataLoader}, configured with prefetching and parallel processing to ensure efficient batch retrieval during training.

\subsection{Dataset and Preprocessing}

\subsubsection{Dataset Characteristics}
\begin{figure}[htbp]
\centerline{\includegraphics[width=0.7\linewidth]{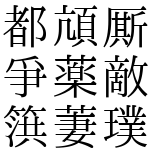}}
\caption{A $3 \times 3$ grid of sample Kanji characters from the dataset \cite{UnlabeledKanji}.}
\label{fig:kanji_samples}
\end{figure}

The study utilizes the Kanji Dataset, a collection of 10,314 black-and-white character images sourced from Kaggle \cite{UnlabeledKanji}. Each image has a resolution of $48 \times 48$ pixels and is rendered in a standardized Mincho-style font. This choice of a synthetic, typographically consistent dataset over handwritten samples was deliberate; it controls for artistic variation and ensures clean topological features. This is a crucial decision for a study focused on comparing the architectural merits of generative models rather than their ability to replicate handwriting. Figure \ref{fig:kanji_samples} displays a sample of characters from the dataset, illustrating their structural complexity.

\subsubsection{Preprocessing and Data Loading}
A standardized preprocessing pipeline was applied before training. First, images were upscaled to $64 \times 64$ pixels using bilinear interpolation to balance detail preservation with computational load. Next, they were converted to single-channel ($1 \times 64 \times 64$) PyTorch tensors. Finally, pixel values were normalized based on model requirements: to a $[0, 1]$ range for the VAE and a $[-1, 1]$ range (using $\mu=0.5, \sigma=0.5$) for the GAN and DDPM.

The models were trained using a batch size of 128 for the VAE and GAN, and 32 for the DDPM due to its higher memory requirements. The data loader was configured with four worker processes, pinned memory, and a prefetch factor of two to ensure efficient GPU utilization. Notably, no data augmentation techniques such as rotation or flipping were applied, as the goal was to preserve the precise structural integrity of the characters for pure generation.

\subsection{Model Architectures and Training}
\subsubsection{Experimental Design}
Three generative approaches were implemented and compared: a Variational AutoEncoder (VAE), a Generative Adversarial Network (GAN), and a Denoising Diffusion Probabilistic Model (DDPM). For each architecture, critical hyperparameters were systematically tested, including latent dimension size (VAE), loss functions (GAN), and noise schedules (DDPM). Model selection was guided by the need for computational efficiency on the available hardware. All models were trained for 100 epochs, after which they generated 10,368 images to ensure a sufficient sample size for accurate Fr\'echet Inception Distance (FID) score calculation.

\subsubsection{Variational Autoencoder (VAE)}
The first model evaluated was a VAE featuring a symmetric residual architecture for its encoder and decoder. Each component was composed of seven residual blocks that progressively adjusted the number of convolutional channels from 64 up to 512 and back down to 64. A standard residual block contained two 3x3 convolutional layers, batch normalization, and ReLU activations, along with a shortcut connection to improve gradient flow. The encoder performed downsampling using convolutional strides of 2, while the decoder used transposed convolutions with corresponding strides for upsampling.

To investigate the impact of representational capacity, the VAE was tested with latent space dimensions of 64, 128, and 256. The model was trained by minimizing a loss function comprising a reconstruction term and a regularization term, as shown in Equation (1):
\begin{equation}
    \mathcal{L}_{\text{VAE}} = \mathbb{E}[\log p(x|\hat{x})] + D_{KL}(q(z|x) || p(z))
\end{equation}
where the first term is the binary cross-entropy (BCE) reconstruction loss and the second is the Kullback-Leibler (KL) divergence, which regularizes the latent space to follow a standard normal distribution \cite{kingma2022autoencodingvariationalbayes}.

Training was performed using the Adam optimizer ($\beta_1=0.9, \beta_2=0.999$) and employed bf16 mixed-precision. A one-cycle learning rate schedule was used, starting at $4 \times 10^{-5}$, increasing to a maximum of $1 \times 10^{-3}$ over 30 epochs, and finally annealing down to $1 \times 10^{-7}$.

\subsubsection{Generative Adversarial Network (GAN)}
The GAN's architecture consisted of a generator and a discriminator, both built with residual blocks and initialized using Xavier uniform initialization. The \textbf{generator} was composed of five residual blocks that progressively decreased the channel count from 512 to 64. Each block contained two 3x3 convolutional layers, batch normalization, and ReLU activations, with transposed convolutions used for upsampling. A final Tanh activation layer scaled the output pixels to the required $[-1, 1]$ range.

The \textbf{discriminator} comprised four residual blocks that increased the channel count from 32 to 512. It used standard convolutions with a stride of 2 for downsampling and LeakyReLU activation functions (negative slope of 0.2). A final linear layer produced a single logit as its output.

To identify the most effective objective for training, three distinct \textbf{loss function} formulations were evaluated. The first was the standard GAN objective using Binary Cross-Entropy (BCE) loss. The second was the Wasserstein GAN with Gradient Penalty (WGAN-GP) loss, which minimizes the Wasserstein distance and uses a gradient penalty term to improve training stability \cite{arjovsky2017wassersteingan}. The third was a hinge loss formulation, another common variant for stabilizing GAN training \cite{lim2017geometricgan}.

The \textbf{training protocol} used separate Adam optimizers for the generator ($\beta_1=0.0, \beta_2=0.99$) and discriminator ($\beta_1=0.9, \beta_2=0.999$), with a constant learning rate of $1 \times 10^{-4}$ for both. To balance the training dynamics, an adaptive discriminator update ratio was implemented: the discriminator was updated five times for every generator update. However, this inner loop would break prematurely if the discriminator's accuracy on a batch exceeded 90\%, preventing it from overpowering the generator and speeding up training. All GAN models were trained using full `float32' precision due to the sensitivity of the training process.

\subsubsection{Denoising Diffusion Probabilistic Model (DDPM)}
The third approach was a DDPM implemented with a U-Net architecture \cite{ronneberger2015unetconvolutionalnetworksbiomedical}. The network's structure included downsampling, bottleneck, and upsampling paths, each containing three custom residual blocks. Downsampling was performed using 2x2 max pooling, while upsampling utilized 2x bilinear interpolation with corner alignment. The main paths were preceded and followed by single convolutional layers.

Each residual block was composed of three 3x3 convolutional layers, 4-group group normalization, and ReLU activations. To incorporate the diffusion time step, sinusoidal position embeddings were processed by a linear layer and added to the output of the convolutional layers, along with the original residual shortcut connection. Filter dimensions increased from 64 to 256 in the downsampling path, rose to 512 in the bottleneck, and decreased from 512 to 64 in the upsampling path. Skip connections concatenated feature maps from the downsampling path to corresponding blocks in the upsampling path. 

Two noise schedules were evaluated over 1024 time steps: a linear schedule increasing from $1 \times 10^{-4}$ to $2 \times 10^{-2}$ \cite{ho2020denoisingdiffusionprobabilisticmodels}, and a cosine schedule \cite{nichol2021improveddenoisingdiffusionprobabilistic}. The model was trained to predict the noise added at a given time step $t$ by minimizing the simplified loss objective \cite{ho2020denoisingdiffusionprobabilisticmodels}:
\begin{equation}
    L_{\text{simple}} = \mathbb{E}_{t,x_0,\epsilon}\big[\|\epsilon-\epsilon_\theta(x_t,t)\|^2\big]
\end{equation}
where $\epsilon$ is the true noise and $\epsilon_\theta$ is the noise predicted by the model. Training used the Adam optimizer ($\beta_1=0.9, \beta_2=0.999$) with bf16 mixed-precision and a cosine annealing learning rate schedule that started at $5 \times 10^{-5}$ and decayed to zero.

\subsection{Evaluation Metrics}
The performance of each model was assessed across several key areas. Generative quality was quantified using the \textbf{Fr\'echet Inception Distance} (FID) score. Computational efficiency was evaluated by measuring \textbf{training time}, \textbf{sampling time} (to generate 10,368 images), and peak \textbf{VRAM usage}. Finally, model complexity was compared by reporting the total number of trainable \textbf{model parameters}.

\section{Results}

The quantitative results of all experiments are summarized in Tables~\ref{tab:VAE-table}, \ref{tab:GAN-table}, and \ref{tab:DDPM-table}. The data reveals that the DDPM models achieved markedly superior performance in terms of Fr\'echet Inception Distance (FID). The top-performing DDPM configuration (linear beta schedule) yielded an FID score of \textbf{26.2}, significantly outperforming the other architectures. For comparison, the best GAN (BCE loss) and VAE (latent dimension 64) achieved considerably higher (i.e., worse) FID scores of \textbf{74.6} and \textbf{79.9}, respectively.

A clear trade-off emerges when analyzing computational costs. The DDPM's high-fidelity generation required an extremely slow sampling phase, taking over 100 minutes to generate the evaluation set. This is more than 2,000 times slower than the GAN, which required only 3 seconds. In terms of training efficiency, the VAEs were the fastest (~14 minutes). Conversely, the GANs, despite having the fewest parameters (8.7M), demanded the longest training times, with the WGAN-GP variant being the most computationally expensive (184.1 minutes).

\begin{figure}[tb]
    \centering
    \includegraphics[width=\linewidth]{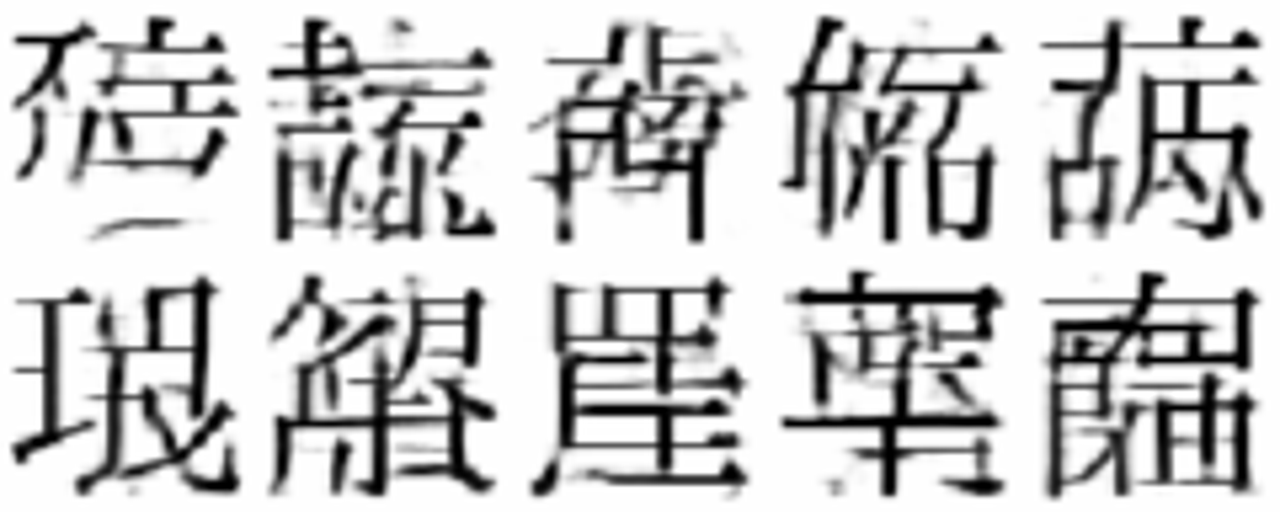}
    \caption{Generated images of VAE with 64 dimensional latent space}
    \label{fig:VAE64}
\end{figure}

\begin{figure}[tb]
    \centering
    \includegraphics[width=\linewidth]{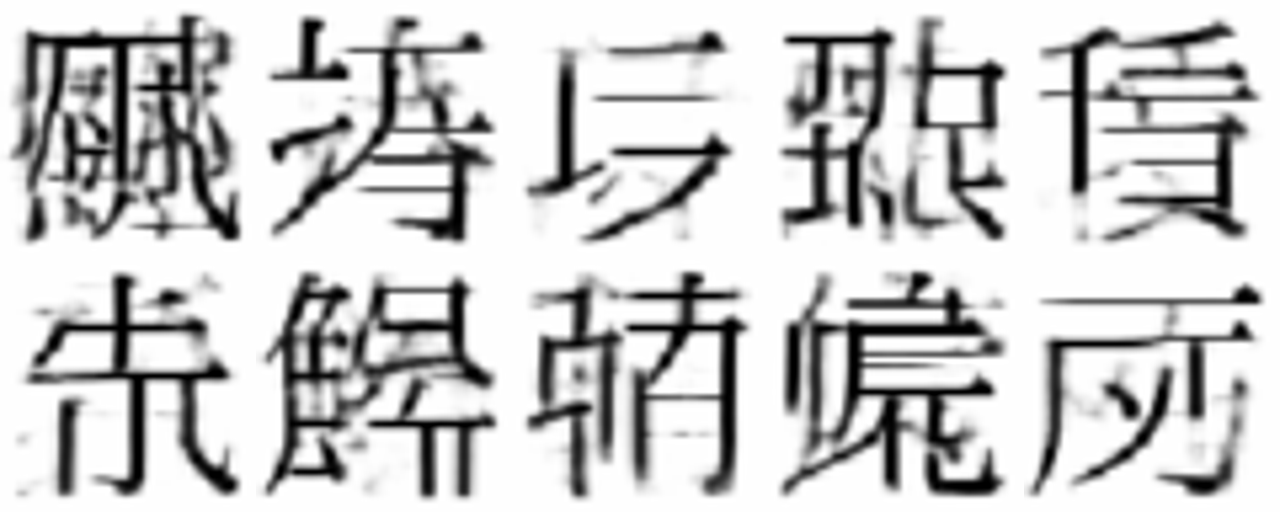}
    \caption{Generated images of VAE with 128 dimensional latent space}
    \label{fig:VAE128}
\end{figure}

\begin{figure}[tb]
    \centering
    \includegraphics[width=\linewidth]{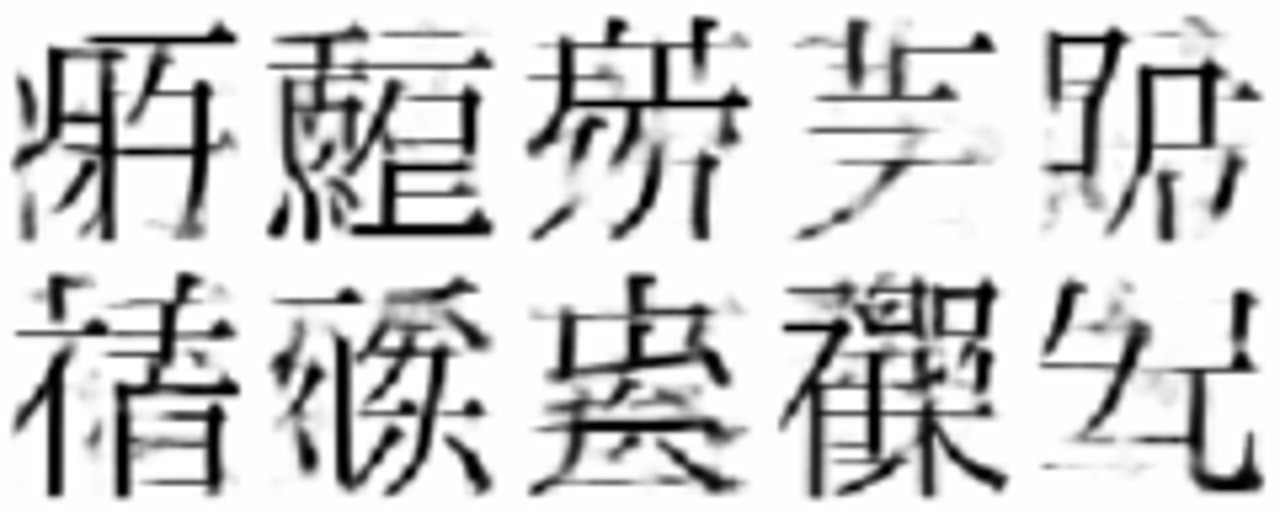}
    \caption{Generated images of VAE with 256 dimensional latent space}
    \label{fig:VAE256}
\end{figure}

\begin{figure}[tb]
    \centering
    \includegraphics[width=\linewidth]{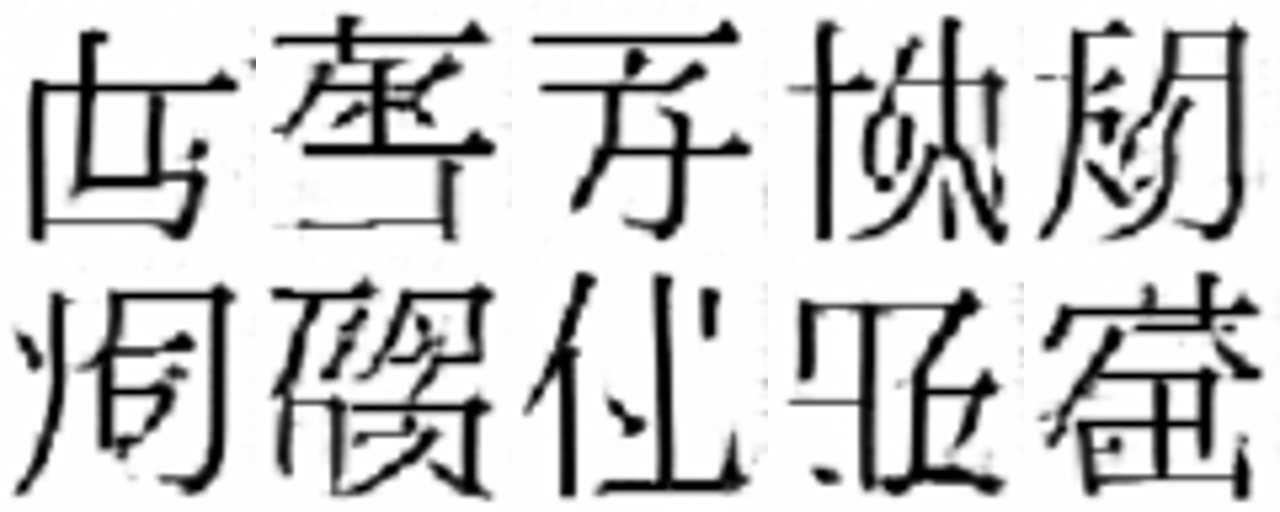}
    \caption{Generated images of GAN using binary cross entropy loss}
    \label{fig:GANB}
\end{figure}

\begin{figure}[tb]
    \centering
    \includegraphics[width=\linewidth]{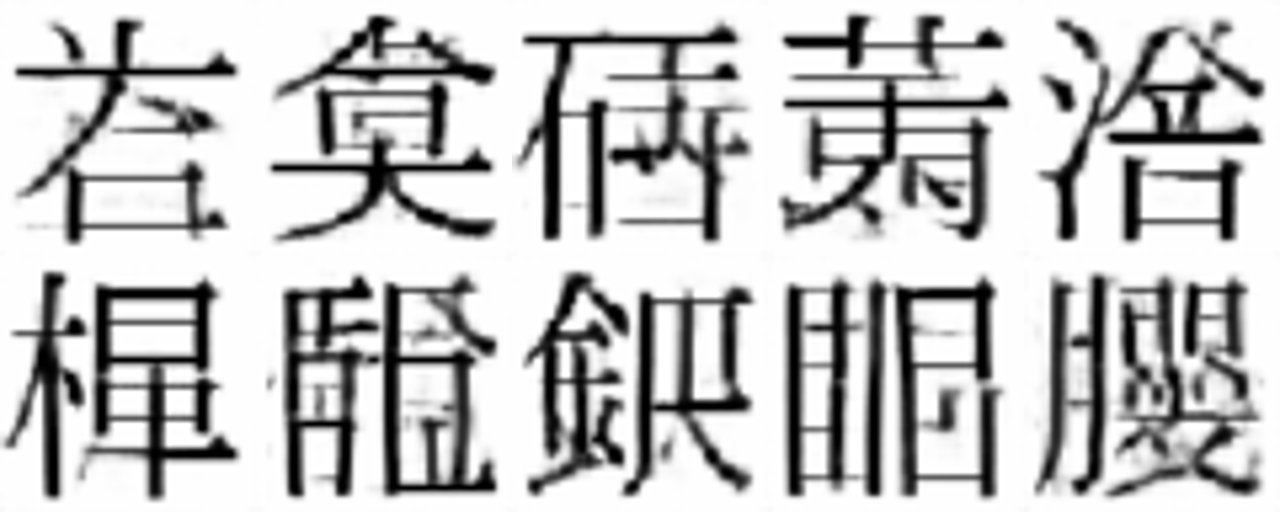}
    \caption{Generated images of GAN using Wasserstein distance loss}
    \label{fig:GANW}
\end{figure}

\begin{figure}[tb]
    \centering
    \includegraphics[width=\linewidth]{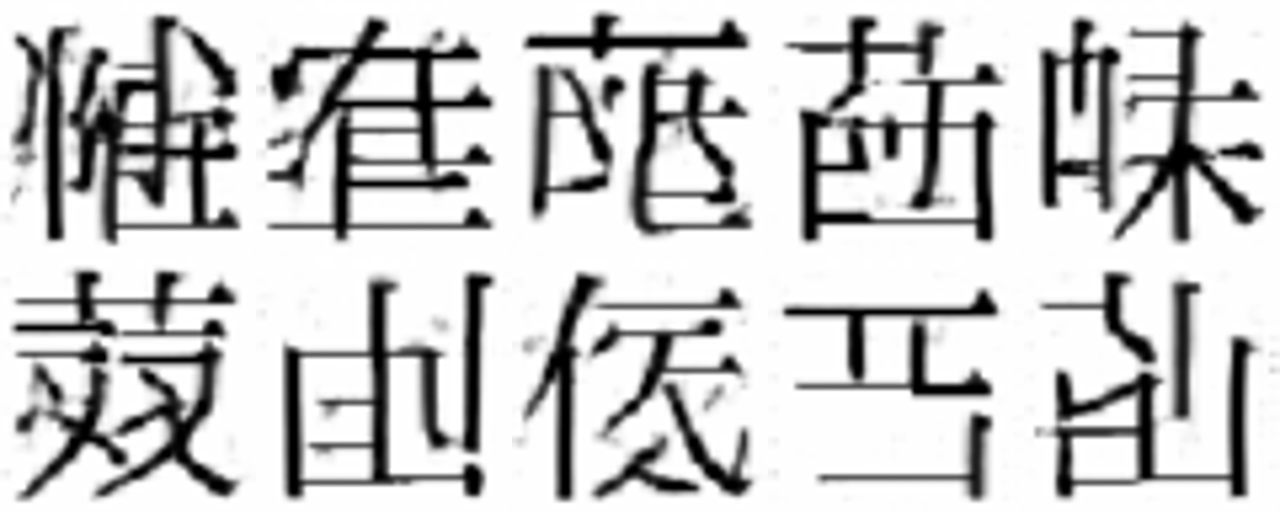}
    \caption{Generated images of GAN using Hinge loss}
    \label{fig:GANH}
\end{figure}

\begin{figure}[tb]
    \centering
    \includegraphics[width=\linewidth]{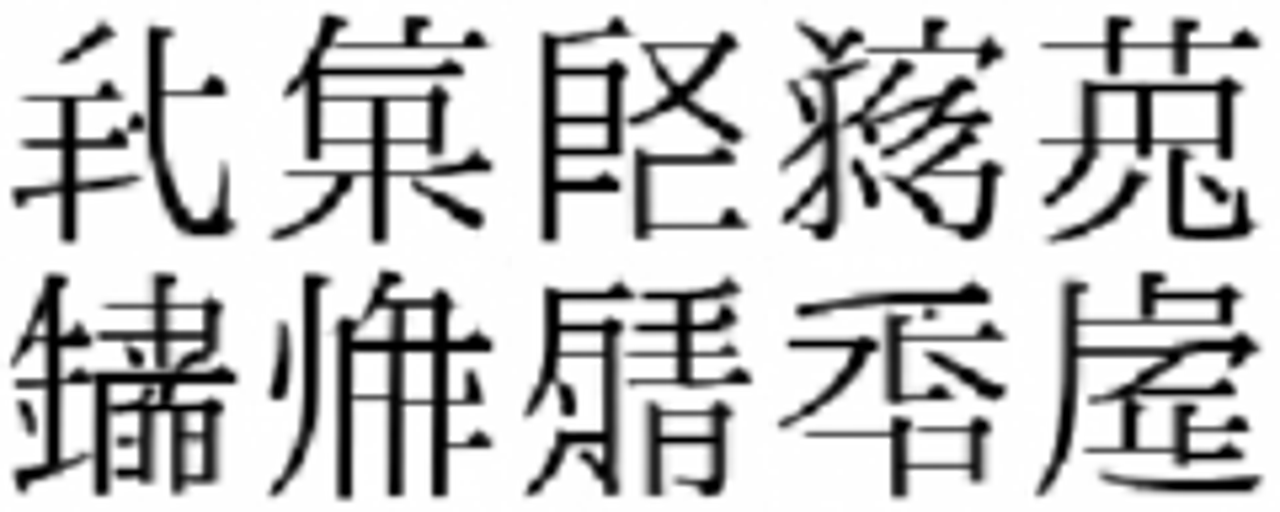}
    \caption{Generated images of DDPM with linear beta}
    \label{fig:DDPM-lin-no-attn}
\end{figure}

\begin{figure}[tb]
    \centering
    \includegraphics[width=\linewidth]{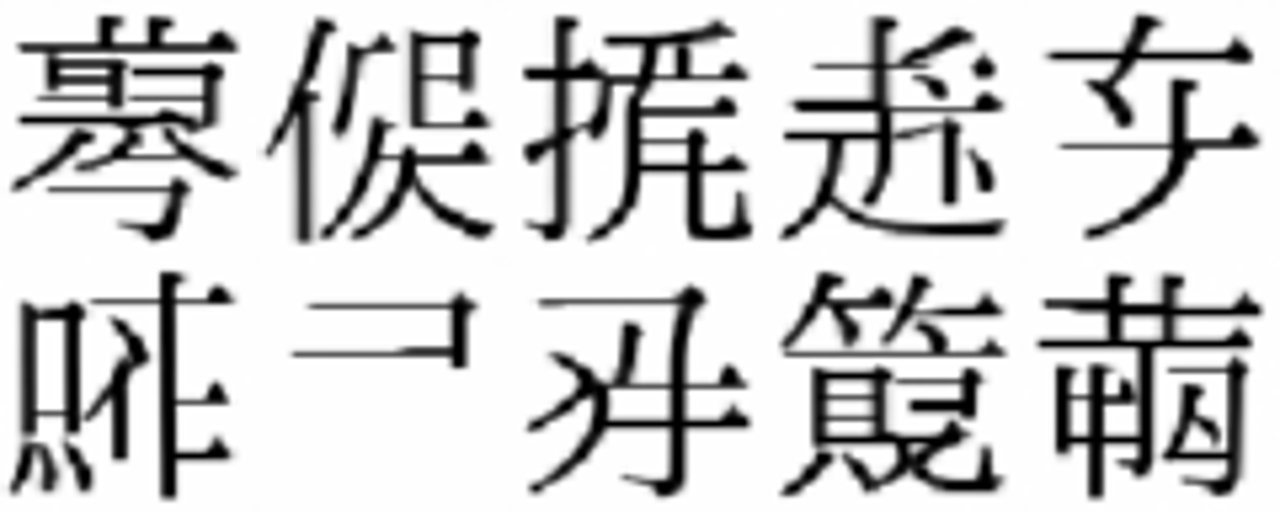}
    \caption{Generated images of DDPM with cosine beta}
    \label{fig:DDPM-cos-no-attn}
\end{figure}

However, a qualitative assessment of the generated images (Figures \ref{fig:VAE64} through \ref{fig:DDPM-cos-no-attn}) highlights a notable discrepancy between the quantitative FID scores and the perceptual quality of the outputs. Images generated by the VAE (Figs. \ref{fig:VAE64}-\ref{fig:VAE256}) are consistently characterized by a soft, blurry appearance that obscures fine structural details. In contrast, the GAN-generated images (Figs. \ref{fig:GANB}-\ref{fig:GANH}), while algorithmically scoring similar to the VAE's, exhibit sharper edges and more defined stroke-like features, rendering them more visually plausible as Kanji characters than their VAE counterparts in this author's qualitative analysis.

\begin{table}[t]
    \centering
    \caption{\textbf{VAE Results}}
    \label{tab:VAE-table}
    \begin{tabular}{l ccc}
        \toprule
        \textbf{Latent space size} & \textbf{64} & \textbf{128} & \textbf{256} \\
        \midrule
        \textbf{FID} & 79.9 & 91.3 & 86.9\\
        \textbf{Training Time} (min) & 14.4 & 14.0 & 14.4 \\
        \textbf{Sampling Time} (sec) & 4 & 4 & 4 \\
        \textbf{VRAM Usage} (MB) & 2458 & 2464 & 2560 \\
        \textbf{Model Parameters} (M) & 22.2 & 23.8 & 26.9 \\
        \textbf{Generated Images} & Fig. \ref{fig:VAE64} & Fig. \ref{fig:VAE128} & Fig. \ref{fig:VAE256}\\
        \bottomrule
    \end{tabular}
\end{table}

\begin{table}[t]
    \centering
    \caption{\textbf{GAN Results}}
    \label{tab:GAN-table}
    \begin{tabular}{l ccc}
        \toprule
        \textbf{Discriminator Loss} & \textbf{BCE} & \textbf{Wasserstein} & \textbf{Hinge} \\
        \midrule
        \textbf{FID} & 74.6 & 78.9 & 78.4 \\
        \textbf{Training Time} (min) & 48.8 & 184.1 & 69.2 \\
        \textbf{Sampling Time} (sec) & 3 & 3 & 3 \\
        \textbf{VRAM Usage} (MB) & 4190 & 6046 & 6046 \\
        \textbf{Model Parameters} (M) & 8.7 & 8.7 & 8.7 \\
        \textbf{Generated Images} & Fig. \ref{fig:GANB} & Fig. \ref{fig:GANW} & Fig. \ref{fig:GANH}\\
        \bottomrule
    \end{tabular}
\end{table}

\begin{table}[b]
    \centering
    \caption{\textbf{DDPM Results}}
    \label{tab:DDPM-table}
    \begin{tabular}{l cc}
        \toprule
        \textbf{Configuration} & \textbf{Linear Beta} & \textbf{Cosine Beta} \\
        \midrule
        \textbf{FID} & 26.2 & 29.1 \\
        \textbf{Training Time} (min) & 31.8 & 31.2 \\
        \textbf{Sampling Time} (min) & 101.7 & 101.7 \\
        \textbf{VRAM Usage} (MB) & 1806 & 1848 \\
        \textbf{Model Parameters} (M) & 23.8 & 23.8 \\
        \textbf{Generated Images} & Fig. \ref{fig:DDPM-lin-no-attn} & Fig. \ref{fig:DDPM-cos-no-attn} \\
        \bottomrule
    \end{tabular}
\end{table}

\section{Discussion}
The results of this study offer a clear comparative analysis of VAEs, GANs, and DDPMs within the specific context of Kanji generation. The superior FID scores of the DDPMs align with established findings that diffusion models excel at high-fidelity, pixel-perfect image synthesis. However, this quality comes at a steep computational price, with sampling times that are orders of magnitude slower than other methods, rendering them impractical for many real-time applications.

Perhaps the most salient finding is the observed paradox between quantitative metrics and qualitative assessment. GANs, despite their challenging training dynamics and susceptibility to mode collapse, produced characters with sharp, defined strokes. In contrast, VAEs, which were the simplest to implement and train, generated blurry, indistinct forms. This suggests that the FID score, while a standard benchmark, may not fully capture the features that are most critical for perceptual quality in structured, line-based data like Kanji. For such datasets, human evaluators appear to prioritize structural integrity and sharpness over the pixel-level distribution matching what FID measures.

These findings validate the proposed \textbf{GANji} framework as an effective ``Hello, World'' for comparative studies in generative AI. The Kanji dataset~\cite{UnlabeledKanji} proved to be an excellent medium, as its inherent structure made the characteristic failure modes of each model—the VAE's blurriness, the GAN's occasional malformations, and the DDPM's residual noise—immediately apparent. The framework successfully highlights the fundamental trade-offs between model complexity, training stability, resource consumption, and the nature of the generated output.

Nonetheless, this study has limitations. The qualitative evaluation of "Kanji-ness" is inherently subjective and benefits from familiarity with the characters. Furthermore, the use of a single, clean font does not test the models' robustness to stylistic variations found in handwritten data. Future work could address these points by developing a quantitative metric for structural coherence, conducting user studies for perceptual scoring, and applying the framework to more diverse or handwritten character sets.

\section{Conclusion}
This paper introduced and evaluated GANji, a lightweight framework for benchmarking generative models on the task of Kanji character generation. Through systematic experiments, it was demonstrated that while DDPMs achieve superior image fidelity as measured by FID, they are hindered by extreme sampling latency. Critically, this work highlighted a significant discrepancy between standard metrics and human perception, as GANs produced more visually coherent characters than VAEs despite having similar FID scores. Ultimately, the GANji framework proves to be a valuable and accessible tool for both educational purposes and preliminary research, effectively revealing the distinct performance profiles and trade-offs of foundational generative architectures.

\printbibliography

\end{document}